**RESEARCH ARTICLE**

**Journal Section**

# Differentiating Geographic Movement Described in Text Documents

## Scott Pezanowski[1]*    |    Alan M. MacEachren[1,2]†    |

## Prasenjit Mitra[1]*

[1]
  Information Sciences and Technology, The Pennsylvania State University, University Park, PA, USA

[2]
  Department of Geography, The Pennsylvania State University, University Park, PA, USA

**Correspondence**
Scott Pezanowski, Information Sciences and Technology, The Pennsylvania State University, University Park, PA, 16802, USA
Email: scottpez@psu.edu

**Present address**
*
  Information Sciences and Technology, Westgate Building, The Pennsylvania State University, University Park, PA, 16802, USA
†
  Department of Geography, Walker Building, The Pennsylvania State University, University Park, PA, 16802, USA

Funding information

Understanding movement described in text documents is important since text descriptions of movement contain a wealth of geographic and contextual information about the movement of people, wildlife, goods, and much more. Our research makes several contributions to improve our understanding of movement descriptions in text. First, we show how interpreting geographic movement described in text is challenging because of general spatial terms, linguistic constructions that make the thing(s) moving unclear, and many types of temporal references and groupings, among others. Next, as a step to overcome these challenges, we report on an experiment with human subjects through which we identify multiple important characteristics of movement descriptions (found in text) that humans use to differentiate one movement description from another. Based on our empirical results, we provide recommendations for computational analysis using movement described in text documents. Our findings contribute towards an improved understanding of the important characteristics of the underused information about geographic movement that is in the form of text descriptions.

K E Y W O R D S
geographic movement, natural language processing, text mining, geographic information retrieval, network analysis, unsupervised clustering, information retrieval

# 1 | INTRODUCTION

Geographic movement information in text-based sources is an underused resource but has great potential to provide valuable knowledge contributions about the movement of things throughout the World. Existing work has been done to detect movement statements automatically (Pezanowski and Mitra, 2020b), and to detect (and interpret) more specific types of movement descriptions like route directions (Jaiswal et al., 2010; Drymonas and Pfoser, 2010). This prior work is a step toward making such data more readily available. Here, we present results of empirical research to understand the characteristics that differentiate movement statements from each other. One key motivation for characterizing differences in movement statements is to provide the base for subsequent methods to take advantage of the wealth of contextual information intertwined with the movement description in text. This contextual information is an advantage for monitoring and understanding movement that movement statements have over Global Positioning System (GPS) trajectories. Contextual information is essential in knowing why the movement is occurring. But, a first step toward leveraging this contextual information from movement statements is to determine features that humans use to distinguish among movement statements. Understanding these features will help put necessary constraints on subsequent computational strategies to capture movement context from text. We also anticipate that an understanding of characteristics that differentiate movement statements will provide valuable input to methods that automatically categorize movement statements, such as those about specific animal migrations, human migrations, or the movement of illegal trafficking of goods, wildlife, or humans.

Geographic movement data analysis (commonly referred to as trajectory analysis since it is dominated by research using sensor-captured trajectories as data) has an extensive body of research. This research has advanced the understanding of disease spread, animal migrations, urban planning, crowd control, and automobile traffic management. Most of this research relies on analyzing trajectories that have precise and frequent geographic coordinates and time stamps acquired through GPS. Although very successful in its own right, this research often does not directly access valuable contextual information about why the movement is occurring. In trajectories, contextual information is either not available, is general information on the thing moving, or is information that needs to be acquired using advanced techniques or manual efforts like field observation or domain expert analysis of the trajectory data. Some research has begun to address this deficiency by semantically connecting the trajectory data with contextual information from other sources (Parent et al., 2013). For example, suppose you use advanced spatial analysis and aggregation techniques to identify stops in trajectories and the stop's time. In that case, this stop information can be used as a point-of-interest (POI) and queried against other datasets to find semantically relevant information (Andrienko, Gennady L Andrienko et al., 2013). Another research effort (focused on animal movement) showed that advanced techniques could be used to link trajectories of tiger movement with other semantic information (Ahearn et al., 2017).

As opposed to trajectories, movement described in text often has the advantage of having much more contextual information about why the movement is occurring intertwined naturally with the text's geographic information. The combination of Geographic Information Retrieval (GIR) methods with computational techniques like Natural language processing (NLP), Topic Modeling, and clustering can be used to extract, analyze, and understand topics in text and convert the derived movement information into a format typically used by mapping software. Descriptions of movement do not have the advantages of GPS data of the precise location and time and therefore cannot be substituted for it. However, since movement descriptions do contain place mentions and have the advantage of readily accessible contextual information, they can be used to gain knowledge about the movement to complement GPS trajectories or be used alone for analysis when GPS trajectories are not available because of technical challenges, cost, or privacy concerns.

As an example that provides further motivation for our research, Figure 1 shows both how geographic movement is implicitly described in text and how it is commonly accompanied by valuable contextual information about why the movement is occurring. The statement clearly illustrates a storm over Avalon Beach, New South Wales, Australia, and the person is nearby watching the storm. They travel via automobile along the freeway and stop in the area of Ourimbah to view more storms. Moreover, perhaps most importantly, the statement clarifies why the person moves - to watch and photograph these intense storms. In addition to this contextual information, there is potentially valuable linguistic information like the author's tone, sentiment, enthusiasm, a time reference, details about the storm's nature itself, and the author's reasoning behind his/her movements and stops. Given this information, it would be straightforward to map this movement as shown in the map in the figure with storms near Avalon Beach in the south, the person's driving route along the M1 freeway north, and more storms near Ourimbah. Also, Information Retrieval (IR) techniques can be used to obtain other information from external sources about the geographic area, time, or storms, if needed.

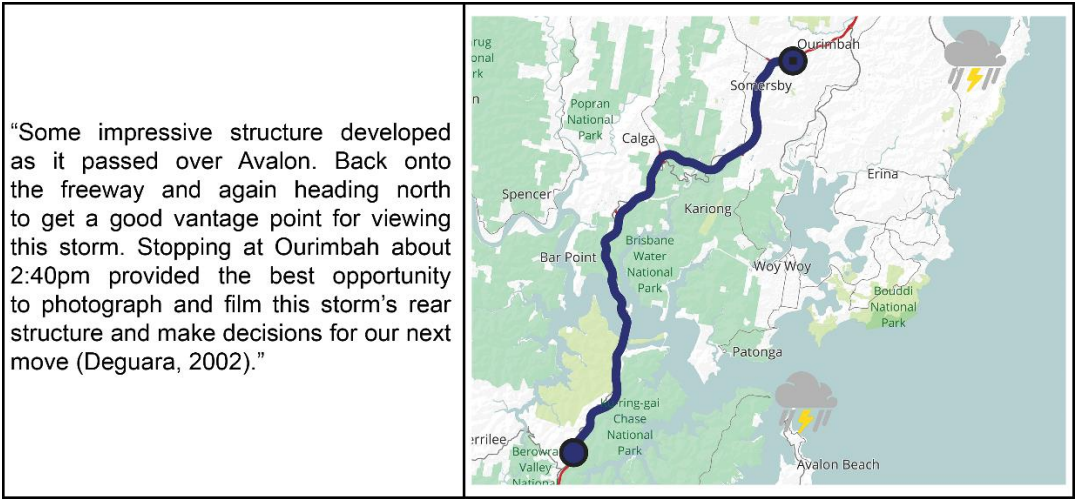

"Some impressive structure developed as it passed over Avalon. Back onto the freeway and again heading north to get a good vantage point for viewing this storm. Stopping at Ourimbah about 2:40pm provided the best opportunity to photograph and film this storm's rear structure and make decisions for our next move (Deguara, 2002)."

**FIGURE 1** An example of a movement description that is representative of the kinds of contextual information that movement descriptions can provide. The map on the right derived from the description produced by the storm chaser illustrates the movement described in the statements to the left, from the storm.

Previous research showed evidence that movement statements are at times difficult to interpret and that people conceptualize the geography described differently (Pezanowski and Mitra, 2020b). In this manuscript, we confirm this and explore some of the reasons why by asking humans to differentiate statements that describe movement and observe and characterize the lack of agreement. Specifically, we posed and answered the research questions: Are there identifiable characteristics of descriptions of geographic movement that people use to differentiate the described movement? If the answer is yes that there are identifiable characteristics, then what are the important characteristics? If the answer is no, there are no identifiable characteristics; then what are the reasons people have trouble differentiating geographic movement descriptions?

To answer these questions, we use a mixed-methods qualitative research approach combining the methods below.

**1.** An experiment using human participants who judge which statements are most different among three.

**2.** A survey after each experiment choice where the participants are asked what characteristics of movement, they used to differentiate statements.

**3.** An extended survey where selected participants of the experiment are asked what characteristics of movement they used overall to differentiate statements and how often for each.

The overall goal of combining these multiple methods is to identify differentiating characteristics, similarities between statements or groups of statements, or important patterns in the statements that other researchers can use as a baseline to advance progress in utilizing movement statements identified within and extracted from text. Similarity among statements is a flexible judgment people make based upon their background and on the immediate situation (Medin et al., 1993). Therefore, to measure similarities between movement statements, we must incorporate human judgment in our research. Based upon our empirical research with human participants from the general public, we not only identify statements that are different from each other, but we identify important characteristics of the movement that tells us why they are different.

The long-term goal of our research is to advance analysis techniques of movement described in text. Text-based data comprises a substantial portion of all data that exists (Schneider, 2016). This vast amount means computational techniques must assist humans in the analysis. In summary, much of the understanding of text is based on human judgments; however, the volume of text existing on any particular topic is typically far too large for a human to comprehend realistically. Therefore, computational methods are needed. In the research reported here, we carried out human subject experiments to identify key characteristics of movement statements. The results of our research can provide input to subsequent research on computational techniques designed to uncover patterns in movement statements on a large scale. Our research findings provide specific recommendations for computational techniques to further analyze movement described in text to make this underused resource more accessible.

## 2 | LITERATURE REVIEW

This literature review begins by highlighting key research in spatial cognition that explains why movement statements are challenging to utilize and why a characterization of their differences is needed. Then, we cite research that shows how researchers have recognized the value of adding contextual information to trajectories and made initial attempts to add context. This research provides motivation to make better use of movement statements since context is more readily available than trajectories. Next, we review research that has characterized trajectories, similar to our intention with movement statements. This research also provides motivation for an improved understanding of movement statements. The research we report in this manuscript will begin to fill a research gap through an extensive characterization of movement statements using both human judgment and analysis of those judgments. Throughout the remaining sections, we include references to necessary related research about the methods used.

Spatial cognition is a long-standing active area of research focused on how people conceptualize entities and relations in the World (Montello, 2015). People conceptualize geography and geographic concepts in many ways depending on their environment, education, experiences, and natural way of thinking. Along with spatial cognition differences, people do have common understandings, for example, with how they perceive geographic scale (Montello, 1993). Ambiguity and differing interpretations of geographic phenomena are common in geographic statements generally. Previous research to create a corpus of movement statements, illustrated this ambiguity (Pezanowski and Mitra, 2020b). The difficulty in interpreting movement statements and resulting labor-intensive labeling encountered in that corpus-building research necessitated machine learning model predictions to enlarge the corpus to a useful size. After the effort to detect movement statements, making the detected statements more valuable requires a deeper understanding of the statements' differentiating characteristics. As spatial cognition has taught us, humans do have common understandings of geographic concepts (Hirtle, 2019), and we propose that this also applies to movement statements' characteristics.

Although movement statements are an underutilized resource, research with trajectories has an extensive body of work and is a growing field (Miller et al., 2019). Existing research with trajectories has improved the understanding of human spatial behavior and activity patterns (Liu et al., 2010; Zhao et al., 2021), wildlife movement (Laube and Purves, 2011; Williams et al., 2020), and the distribution of goods (Andrienko and Andrienko, 2021; Wasesa et al., 2017; Hadavi et al., 2019), to name a few. Providing richer context (contained within statements about movement) can enhance the success of geographic movement research. Andrienko, Gennady L Andrienko et al. (2013); Ahearn et al. (2017) attempted to address the lack of context in trajectories by creating automated ways to add contextual information to trajectories. Siła-Nowicka et al. (2016) notes this deficiency and shows how the contextual information of known gender and age of human subjects moving can be used to interpret movement data better. In a second example, topographic differences (mountainous vs. lowland) found in linked data can explain why the same breed of tigers in Thailand have very different movement patterns in different places (Ahearn et al., 2017). A few other research efforts connected trajectories with other contextual data that is in the same geographic location (Chen et al., 2019), identified important places in movement data to divide related descriptions of movement by different travel modes (Sester et al., 2012), used computational methods to derive pedestrian behaviors (Torrens et al., 2011), identified different motion types (Weiming Hu et al., 2004), and identified modes of travel (Zheng et al., 2010). This research to add contextual information to trajectory data analysis is in its infancy but shows the potential of such information to advance research in the field. More importantly, these efforts show that having context information common in movement statements readily accessible can provide valuable knowledge if those movement statements were made more accessible through the thorough characterization of their differences we have done.

The research noted above provides examples showing specific advances in and applications of movement analysis with trajectories. Other research has provided frameworks for conceptualizing key attributes of trajectory data. Andrienko et al. (2008) defined basic concepts of movement data, Dodge et al. (2008) created a taxonomy of movement patterns, and Alamri et al. (2014) created a taxonomy of moving object queries. These research efforts provide a baseline for analysis for other research that utilizes trajectories by characterizing trajectories and movement data with these attributes and taxonomies. These frameworks have been used to organize research analysis into traffic interchange patterns (Zeng et al., 2013), detection of anomalies in traffic (Orellana et al., 2009), identifying key segments of trajectories (Ferrero et al., 2018), and construct database queries specific to moving objects (Zhang et al., 2016), to name just a few. Although these frameworks are for trajectories, they overlap with movement statements in that they are about geographic movement. Thus, we used them as a starting point for developing methods to characterize differences in movement statements.

Although all the existing research in this area has supported advancements in movement analysis, the prior studies focus on movement data where location and time of movements are precise. Researchers have not addressed taxonomies of movement information derived from text sources where that information is imprecise. But the success of these frameworks for trajectories provides motivation for our research into a similar baseline for movement statements. The challenges to using movement described in text - ambiguousness of place mentions, the inherent flexible interpretation of written text, and movement that is often generalized both in space, time, and things moving - can be better addressed given a thorough characterization of their differences. The research reported here focuses on identifying characteristics important to differentiating movement statements and, thus, is a step toward developing a formal taxonomy of movement statements that complement the ones developed for trajectories.

## 3 | RESEARCH PLAN AND EXPERIMENT DESIGN

Our primary research data gathering technique is an experiment using Amazon Mechanical Turk (MTurk). We present a diverse set of statements that describe movement to human participants and ask them to judge differences. Although there is some debate about the quality of MTurk experiment data, research has shown that when the experiment is set up as a simple task and MTurk options are used to restrict the participants to those who do quality work, quality results are likely (Klippel et al., 2013).

More specifically, using statements that describe movement, we first perform a pilot experiment where we identified the four most important characteristics to focus on. The pilot experiment also identified some problems in our experimental setup that we corrected in the full experiment. Then, we designed the main experiment as described below and summarized in Figure 2. The main experiment used a *triad comparison of differences* with human participants. These participants are asked to judge which statement among the three statements is most different. This task was done with 308 statements that describe geographic movement that were selected from a corpus systematically built from diverse sources on the WWW. This exercise is meant to 1) provide the participants with clear and realistic examples of statements describing geographic movement and 2) prompt them to think about what makes the movement described different. After each triad task is completed, the participant is asked a follow-up question to pick the characteristics from the four characteristics identified in our pilot experiment that they used to decide the most different statement. They can choose from zero characteristics to all four characteristics.

Last, participants who completed ten or more triads (these active participants completed 86% of all triads for the experiment overall) were invited to complete an extended survey. They are asked to provide more details about the characteristics they used to differentiate the statements. This extended survey first asked the participants to rate (for the full set of tasks they completed) the amount they used each of the four primary characteristics of movement on a five-level scale. We decided to focus on these four characteristics, as we did in the short per-triad survey because our pilot experiment survey showed they were used most often to differentiate movement described in statements. Then, participants were asked to choose any additional characteristics they used from an extended list of potential characteristics of statements describing movement. This extended list was based on both a) ideas derived from some of the frameworks/taxonomies of movement developed for trajectories and b) our own initial review of results from our pilot experiment (described below). Our research plan and experiment design are summarized in Figure 2.

In this section, we describe our experimental setup in subsections about 1) the platform and participant selection, 2) selecting, curating, and constructing triads from the statements, 3) the participants' task, platform, and participant selection, 4) the triad comparison task, 5) the short survey given for every triad, 6) the long survey given after completing all tasks, and 7) a summary of differences between the pilot experiment and full experiment. Together as a whole, we use these multiple research methods to identify important distinguishing characteristics of movement statements. These characteristics create a baseline understanding of movement found in text, to further the research field.

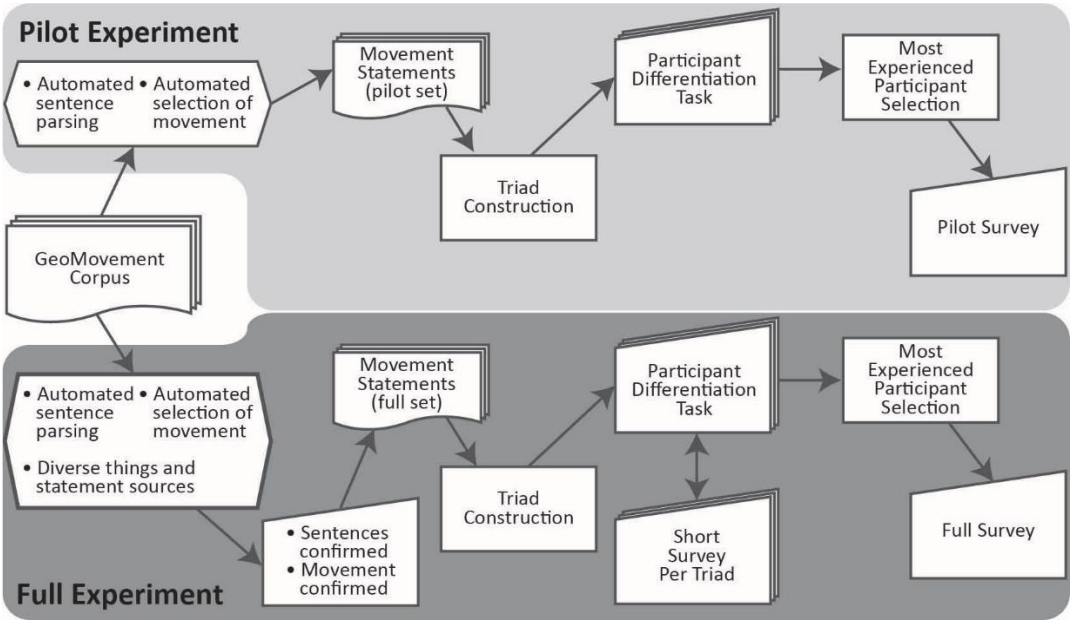

**FIGURE 2** Research plan experimental summary.

## 3.1 Platform and Participant Selection

We used Amazon Mechanical Turk (MTurk) as the platform for judging the similarity of movement statements that were presented in triads. MTurk was used in both the pilot study and the main experiment. MTurk allowed us to utilize a large number of participants anonymously for the experiment MTurk has been used extensively with success as a platform to hire human participants to do research work that does not require expert knowledge (Paolacci et al., 2010; Kittur et al., 2008; Klippel et al., 2013). MTurk participants are known as "workers" and are paid a small amount for short repetitive tasks that can be performed based on easy-to-understand instructions. A typical example of research work that MTurk has had success for is labeling training data for machine learning tasks like speech and language analysis (Callison-Burch and Dredze, 2010) or image recognition (Rashtchian et al., 2010). MTurk options make it possible to restrict workers to ones who have done quality work in the past. We used this option to improve the probability of getting quality answers.

Some researchers have expressed concerns about the quality of MTurk results (Paolacci et al., 2010). However, in addition to the many successful research experiments discussed, it has been shown that results can be even better than a traditional academic research technique of using students as participants in specific tasks (Kees et al., 2017; Moss and Litman, 2020). A second concern raised about using MTurk to acquire participants and run experiments is the potential to exploit workers for low pay. This concern is alleviated by paying workers the equivalent of a minimum wage hourly salary or more based upon the expected time consumption per task. One important advantage of using MTurk (over traditional in-person laboratory experiments) is that the platform makes it easy to obtain a broad representation of English speakers. Our post-task general survey was conducted using the Qualtrics (https://www.qualtrics.com/) survey platform. Figure 3 shows the locations of participants who completed the triad tasks plus the post-task general survey (the participant's locations are provided by the Qualtrics platform). The MTurk workers were paid $0.08 per triad/short survey answer combo, and the full survey participants (who, as noted above, had completed the minimum of ten triad tasks in the experiment) received $1.50 for completing this follow-up survey. Participants were only allowed to access the initial triad tasks if they had an approval rate greater than or equal to 95% on their previous MTurk work.

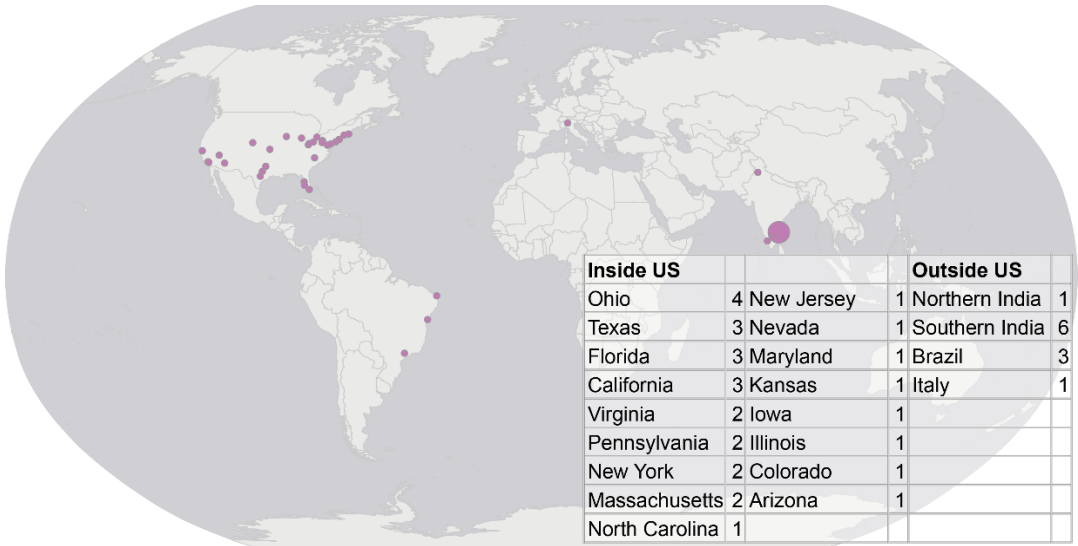

| Inside US | | | | Outside US | |
|---|---|---|---|---|---|
| Ohio | 4 | New Jersey | 1 | Northern India | 1 |
| Texas | 3 | Nevada | 1 | Southern India | 6 |
| Florida | 3 | Maryland | 1 | Brazil | 3 |
| California | 3 | Kansas | 1 | Italy | 1 |
| Virginia | 2 | Iowa | 1 | | |
| Pennsylvania | 2 | Illinois | 1 | | |
| New York | 2 | Colorado | 1 | | |
| Massachusetts | 2 | Arizona | 1 | | |
| North Carolina | 1 | | | | |

**FIGURE 3** The locations of the 41 full survey participants as provided by the survey platform.

## 3.2 | Experimental Data Selection, Curation, and Triad Construction

As mentioned earlier, a random selection was performed to select the statements for the experiment from a corpus of statements predicted to describe geographic movement (Pezanowski and Mitra, 2020a). This corpus was generated by a process involving a) humans searching and labeling movement statements, b) multiple modern techniques to create a machine learning model to predict more movement statements, and c) human corrections of model predictions to produce more accurate models. Extensive details on this process are found in Pezanowski and Mitra (2020b). All statements selected from the corpus were reviewed by hand to ensure they described movement and were complete statements. Any that did not describe movement or were not well-formed sentences were discarded. This manual correction ensured that:

1. The statements did describe geographic movement.
2. All movement statements are full, well-formed sentences.
3. A substantial portion of statements describes people-related movement, and a large portion describes wildlife-related movement for a diverse set.
4. A substantial portion of statements are from news articles, general web pages, and scientific articles, respectively, for diverse writing styles.

The result was 308 total statements. The disadvantage of the hand confirmation step is that the total number of statements for the experiment was limited. All statements were labeled by the type of thing moving and the type of source the statement came from to make sure there was diversity in the things moving and movement described. For the type of thing moving, 185 statements involved people moving, with a few of these being people moving while transporting other things like goods. The other 124 statements were about wildlife moving, with a few about wildlife moving while transporting other things like disease. For the statements' source type, 83 statements were from scientific articles, 149 were from general web pages, and 77 were from news articles. Since there are three statements per triad and each statement was included in three triads, the total number of triads also equaled 308. Each triad task was completed by five different participants producing a total of 1,540 completed triad tasks.

The triad experiment involved placing each movement statement into multiple triads with two different semi-randomly chosen statements and asking multiple human participants to complete the cognitively simple task of judging which of the statements is most different. The semi-random selection for one triad is shown in Algorithm 1, where a random selection is made on those statements that are of similar length, have not been chosen the maximum number of times, and have not been chosen as part of the current triad. This semi-random selection method elicits human judgment in a similarity comparison between statements where participants who complete many triads see a wide range of statements. The two statements that remain after the most different one is picked are considered to be relatively more similar to each other than the one identified as most different. Each statement was included in three different triads with two other semi-randomly chosen statements of similar length. Similar length statements were desired to avoid a scenario such as two lengthy statements being included in a triad with one short statement and the visual difference in length affecting the participants' choice. The specific process for picking statements of similar length used the following steps: 1) randomly choose one statement for the triad that was included in less than three other triads, 2) filter all statements by those whose length is within 200% of the length of the first statement and vice-versa and are a) not the first statement and b) statements not already included with this statement in a triad, 3) if a filter for similar length statements produced less than two statements, increase the percentage of the allowable length between statements by 10%, 4) repeat this process until at least two statements are found, 5) out of these filtered statements, randomly choose two statements to complete the triad. Every time a statement is included in a triad, the three statements are recorded so that the three are not combined again and so that they are included in a total of 3 triads. The process of selecting three statements for one triad is shown in Algorithm 1.

---

**Algorithm 1:** Selecting three movement statements for one triad. **Output:** A triad of three statements

1 first_statement ← select first random statement **where** (**in** < 3 triads);

2 statement_length_ratio ← 200%;

3 new_statement_1;

4 new_statement_2;

5 **repeat**

    6 new_statement_1 ← select random statement **where**

       (**not** first_statement) **and**

       (**not** new_statement_2) **and**

       (**in** < 3 triads) **and**

       (**length** new_statement_1 < **length** first_statement * statement_length_ratio) **and**

       (**length** first_statement < **length** new_statement_1 * statement_length_ratio);

    7 new_statement_2 ← select random statement **where**

       (**not** first_statement) **and**

       (**not** new_statement_1) **and**

       (**in** < 3 triads) **and**

       (**length** new_statement_2 < **length** first_statement * statement_length_ratio) **and**

       (**length** first_statement < **length** new_statement_2 * statement_length_ratio);

    8 statement_length_ratio ← statement_length_ratio + 10%;

9 **until length** *new_statement_1* ***exists*** **and** new_statement_2 ***exists***;

```
10 return (first_statement, new_statement_1, new_statement_2);
```

## 3.3    |  Participant Task to Judge Most Different Statement in Triad

As a first step to identify differentiating characteristics of movement statements, we asked humans to perform a task where they would do exactly this - differentiate statements describing movement. Since differentiating movement statements is a difficult task to do and difficult to conceptualize, it is important to have clear and realistic examples. As outlined above, participants were given three semi-randomly selected statements in each triad comparison. This triad comparison task asks participants to pick the most different movement statement among the three statements. Comparing differences in movement statements is challenging because of the linguistics attributes we mentioned before. Therefore, we decided to use this triad comparison where their task was to judge the most different statement because this task is a very cognitively simple task. Plus, with MTurk, this "most different" task requires just one mouse click, whereas asking participants to pick the two that are most similar would require two clicks.

Figure 4 presents a sample single triad task that three different workers were shown. Their task was to read the three statements (in the left side of Figure 4) and judge which statement represents the most different kind of statement about movement. Each statement they were asked to focus on is shown in larger bold text, while the smaller, lighter text provides the context of the statement. By choosing a statement where the described movement is most different, we can assume that the participants think the other two statements are more similar. One advantage of a task where participants chose the most different statement over one in which they pick the two most similar is that it involves only one click for the participant rather than two, thus making the task more efficient for the work and less prone to error.

The participants were intentionally given minimal instructions about the focus on geographic movement. The specific criteria for judging difference, beyond the instruction to focus on the category of movement statement and not the thing moving, were left up to the participants. Therefore, this task's goal is to let the participants interpret the statements and judge differences as naturally as practical in the context of an experiment. Participants were explicitly told not to focus on the thing moving because this would focus attention on the kind of thing discussed rather than on the nature of geographic movement. Once we opened the experiment, all 1,540 triads were completed within 12 hours.

FIGURE 4 An example of one MTurk task asking the participant to choose which statement represents the most different kind of statement (left) and the follow-up question about their strategy used to make a decision (right).

## 3.4 | Short Survey about the Most Important Characteristics

We conducted two surveys in our full experiment. First, a short survey was given for every triad where the participant was asked to indicate which (if any) of the four main characteristics, listed on the form shown in Figure 4, they used in their decision of the most different statement in that triad. These four choices were based on answers provided to a similar question in our pilot experiment about key characteristics used for differentiating movement statements (more detail about the pilot experiment and what was learned from it are reported in Sections 3.6 and 4.1 below). An example of the short survey question is shown in Figure 4 in the second question at the bottom of the right side of the figure.

## 3.5 | Full Survey on Rigorous Characteristics

As a complement to the task-by-task survey with four choices of factors underlying each choice, MTurk participants who completed ten or more triads were invited to take an extended survey. In this follow-up survey, they were asked details about their overall use of the four characteristics already considered in each post-task survey and about the other characteristics they used. Specifically, they were asked (concerning the full set of tasks they completed) how much they used each of the four main characteristics in judging differences on a five-level scale of1) never, 2) rarely, 3) regularly, 4) frequently, 5) always. The first question was:

*In judging differences in geographic movement statements, please rate the extent which you used each of the characteristics below in deciding which is most different. Select the choice that is closest to representing how often you used each characteristic.*

The four primary characteristics given as answer options for the first question are listed in the top table of Table 3.

Next, the full survey participants were given a list of an additional sixteen possible movement characteristics that we compiled from literature about geographic movement in trajectories. They were asked to choose all characteristics from the list that they used. The second question was:

*In addition to the four characteristics focused on above, in judging differences in geographic movement statements, mark (with a check in the box) each of the following characteristics of the statements that you used to decide which is most different. Mark all that you used, if any, and write any other characteristics you used not listed here in the last box.*

The sixteen additional characteristics to choose from are listed in Table 4. The option for "other" was included, but no participants entered another characteristic. The characteristics were presented to the participants in a random order for each participant. Forty-nine participants were sent invitations, and 41 completed the survey. The participants were sent the survey within 12 hours from when the last triad was complete, and all surveys were completed within 24 hours of the last triad.

In the sections below, the full survey that was given at the end of the initial pilot experiment is labeled as the *pilot survey*, the short survey from the full experiment given for every triad is referred to as the *short survey*, and the full survey for the full experiment given to select participants who completed many tasks is the *full survey*.

## 3.6 | Summary of Pilot Experiment and Full Experiment Differences

As indicated above, prior to the full experiment, we conducted an initial pilot experiment using a similar experimental setup. The objective of the pilot was to refine the experimental methods, and it also was used to identify key characteristics of movement statements to ask participants about in the full experiment. The four differences in the experimental setup for the pilot experiment are that in the pilot 1) full instructions were presented in a closed pane that needed to be expanded to read, 2) there was no short survey, 3) the survey only asked participants to check (on a list) all characteristics they used to make judgments about the most different statement in triads, and 4) a different set of statements were used in which a small number were subsequently determined to not describe movement. While the pilot experiment identified some flaws in the procedure (points 1 and 4) that were fixed in the full experiment, it also produced initial results about the most important characteristics used to judge the similarity of movement statements. Specifically, we developed an initial ranking of the most important characteristics of differentiating movement statements. This ranking was used to select the four characteristics that we focused on in the full experiment.

As noted, the pilot experiment also highlighted some flaws in the experiment setup and identified aspects of the full survey that could be improved. The most important of these is that because the statements used were parsed and predicted to be movement by computer algorithms with known limitations, some were not well-formed sentences, and some did not actually describe movement. These flaws prompted us to incorporate a manual check during our process of selecting sample movement statements from the large corpus; statements were selected randomly, but any that were not complete sentences or that did not refer to geographic movement were removed.

## 4 | EXPERIMENT RESULTS: IMPORTANT CHARACTERISTICS OF MOVEMENT STATEMENTS

In our experiments, participants viewed and considered geographic movement from a diverse set of statements and then completed survey questions about what characteristics of the statements were important to them in differentiating the movement described. These experiments and surveys showed that there are some characteristics that are most important and that individuals in our diverse set of participants naturally interpret the movement differently.

In this section, first, we interpret the results from our pilot experiment's survey where the participants identified primary movement characteristics that are most important to them. Next, we perform an extensive investigation of characteristics of the statements using our experiment full survey. After analyzing all characteristics, we show that some participants tend to use spatial characteristics to differentiate movement statements while others do not. Lastly, we compare participant short survey answers completed while viewing statements to the full survey answers completed by the participants while considering the statements in hindsight.

### 4.1 | Identifying Four Primary Characteristics of Movement Statements

Our pilot experiment was conducted with participants completing triad tasks of example movement statements, judging differences in the movement described. Out of all participants in the triad task, we considered those who completed ten or more triads to be experienced with the movement statements. These participants were sent invitations for the pilot survey. The pilot survey results are shown in Table 1, with the number of participants (out of 35 who completed the survey) who chose each characteristic as important. The most used characteristic was *types of places similar – nation, city, ocean, forest*, which was selected by 77% of the participants. *Size of the area covered – World, continent, country, state, city* was selected by 71%. The most common temporal characteristic was *Movement acceleration,* chosen by 71% of participants. *Movement duration – long, short* was chosen by fewer participants, with 63% choosing it. The acceleration would be a difficult concept to visualize from these types of written text. Given our extensive experience with the statements, we feel it is likely that the participants were confusing acceleration with duration.

Our pilot survey showed other important characteristics like the *types of places similar – nation, city, ocean, forest,* and movement mentioning the *same or similar places*. Both characteristics were also designated as important in the full survey. The value of this finding will be discussed more in Section 5 since some GIR geoparsing systems have successfully used the types of places to improve geographic recognition of place mentions in text. Our experience with the statements makes us confident that both spatial characteristics are important.

Based on what we learned in the pilot, we designed a modified version of the survey to invite selected participants to complete it. In this survey, we decided to both repeat the survey to confirm or refute the results and choose four characteristics to focus on in a Likert-style question. Since *Size of the area covered – World, continent, country, state, city* was an important spatial characteristic, we included this characteristic as a spatial attribute in the Likert-style question. For an important temporal characteristic, as noted above, we had concerns about the high ranking of *movement acceleration rate.* We chose to include *movement duration – long, short* in the Likert-style question since it ranked high, and we also felt it is a characteristic that is reasonable to visualize.

A second spatial characteristic we chose to add for the full survey and include in the Likert-style question is *general movement vs. a specific route*. Our complete list of characteristics in the Likert-style questions and the longer list were derived from frameworks related to trajectory research. This characteristic was not directly included in these past frameworks, most likely because, as mentioned in the introduction, trajectories are commonly recorded with precise sensors. Forms of general movement would likely be most important to trajectory research once the analysis is performed to aggregate movement into overall patterns, such as analyses on big data movement patterns (Graser et al, 2021). Therefore, although this characteristic was not included in earlier frameworks, we saw the potential importance of this characteristic in our own analysis of movement statements and in recent research to aggregate big movement data. As a final characteristic to focus on in a Likert-style question, we chose the *type of thing moving*. In the triad tasks for the main experiment, we instructed the participants not to focus on the type of thing moving (since we wanted them to concentrate on the way movement was described rather than the thing moving). But we hypothesize that it is a natural way to differentiate movement statements by kind of thing moving. So, although we did not include it in the pilot experiment as a survey choice, we decided it is important enough to include it in the full experiment.

**TABLE 1**  The number of times each characteristic was chosen as important in differentiating movement in the pilot survey by the 35 participants.

| Characteristic | # of times chosen |
|---|---|
| types of places similar - nation, city, ocean, forest | 27 |
| same or similar places | 26 |
| movement acceleration rate | 25 |
| distance traveled | 25 |
| size of area covered - World, continent, country, state, city | 25 |
| individual or groups of things moving | 25 |
| movement repeats at intervals - monthly, yearly | 25 |
| things leave and return to the same place | 24 |
| movement caused by other incidents | 22 |
| movement duration - long, short | 22 |
| direction traveled | 22 |
| things moving that intersect with other things | 22 |
| number of stops | 20 |
| time ordered sequence of movements | 19 |
| mode of transportation | 19 |
| movement speed - fast, slow | 14 |
| number of turns | 12 |
| continuous or non-continuous movement | 9 |
| other | 6 |

## 4.2 | Extensive Investigation of Characteristics of Movement Statements

For the full triad experiment, there were 142 participants total who completed at least one triad. Of these 142 participants, 54 completed only one triad. Eight participants completed more than 40 triads, with a maximum of 84 for one participant. There are two pieces of evidence that suggest the participants took the task seriously. The first is the high dropout rate. Since we limited participants to those with high ratings in MTurk, our interpretation is that some participants who saw that the movement statements are difficult to differentiate chose not to continue instead of continuing and doing shoddy work. These participants could have continued and quickly clicked through answers just to receive payment but chose to instead stop. Research has shown that MTurk workers' desire to maintain a high approval rate is strong so that they can continue working (Hauser et al, 2019). Second, MTurk provides the amount of time spent on each triad/short survey combo by recording when the participant accepted the task and when they submitted it. If a participant was simply quickly choosing an answer so they can get paid and not trying to complete the task, from our own trials, we estimate they can consistently do this in less than five seconds. Out of the 1,540 total triads, only 29 triads (2%) were completed in less than five seconds. Therefore, those participants who continued took sufficient time per task to suggest that they were being conscientious.

As shown in Table 2, complete agreement of all three workers on a triad was uncommon, and the frequency with which participants agreed by a majority on the most different statement was similar to the frequency with which they did not. Lack of agreement is also reflected in a quite low Fleiss's Kappa of 0.013. The lack of agreement between participants is representative of both the movement statements themselves and how people interpret the statements. It is unlikely the lack of agreement resulted from participants making random choices because (as noted above) we allowed only participants with a high acceptance rate on their previous work, the dropout rate was high, and most of the time, participants took a reasonable amount of time per task. In general, people can easily interpret written text differently. Also, movement described in text can have vague or ambiguous place references, vague time references, and contextual clues that need to be interpreted (Pezanowski and Mitra 2020b). People can interpret the movement statements differently based on their education level, geographic location, and other background knowledge. Despite the lack of agreement in the task of determining the statement that is most different, it serves the purpose to prompt participants to think about the movement described and what characteristics of the movement described are important to them. We considered participants who completed at least ten triads to be experienced participants and selected them for our full survey. A total of 142 participants completed at least one triad.

**TABLE 2** The vote breakdown for agreements between statements in a triad. A vote of 5-0-0 means complete agreement on the most different statement, while 2-2-1 indicates the greatest disagreement in which two participants chose one statement, two chose another, and one chose the third.

| Votes | Count |
|-------|-------|
| 5-0-0 | 10 |
| 4-1-0 | 58 |
| 3-1-1 | 79 |
| 3-2-0 | 76 |
| 2-2-1 | 85 |

There were 48 participants (34% of the total) who completed at least ten triads, and their full survey responses and corresponding MTurk short survey responses were selected for further analysis in the sections below. Interestingly, although the experienced participants were only 34% of the total, they completed 86% of the triads. Within 12 hours after all triads were complete, we sent the selected participants their invitation to participate in the extended full survey. Of these 48 invitations, 41 participants completed the extended full survey.

For each of the four primary characteristics that we asked participants about, we calculated the mean and standard deviation for Likert scale answers from 1, meaning the characteristic was never used by the participant to 5, meaning the characteristics were always used. Table 3 shows that out of the four primary characteristics that participants were asked to consider, *time duration* had the greatest mean (meaning that the participants used it the most on average) and the lowest standard deviation (meaning that the participants agreed the most). The *area covered* was used less than the other two characteristics, which are the type of thing moving and *general movement vs. a specific path*. From the results below, it seems people may be confusing the *size of the area covered* with the area's location and perhaps the distance traveled. Figure 5 shows the distribution of answers to the Likert style questions for the four primary characteristics. This figure highlights the popularity of time duration and the polarity of the geographic characteristic, *size of area covered*.

**TABLE 3** The four characteristics separated for the Likert scale question in the top table show time duration with the greatest mean and the size of the area covered as the greatest standard deviation.

| Characteristic | Mean | Standard Deviation |
|---|---|---|
| Movement duration - long, short | 3.36 | 1.23 |
| Size of area covered by movement - World, continent, country, state, city | 2.91 | 1.36 |
| Type of thing moving | 3.11 | 1.32 |
| General movement versus specific route | 3.11 | 1.32 |

Figure 5 shows the breakdown of how frequently the participants used each of the four main characteristics (also highlighted in Figure 6 below when participants were aggregated into bins). A group of participants relied on the *time duration* a lot, and another group ignored it most of the time. The same can be said for the *area covered* by the movement. These two characteristics were the most polarizing compared to the other two, which were more evenly distributed. Despite the instructions not to focus on the thing moving, a large portion of participants still said this factored into their decision. We hypothesize that this is either because it is simply a natural way to distinguish descriptions of movement or because things that are the same are more likely to have similar movement patterns as compared to different things.

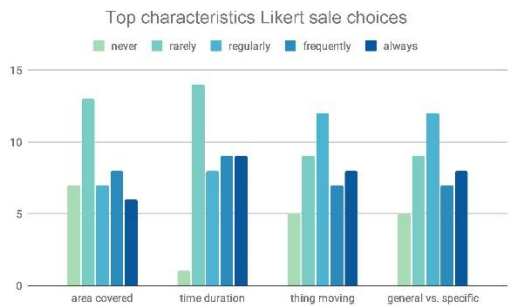

**FIGURE 5** The distribution of Likert style question answers to the four primary characteristics.

Among the long list of other characteristics (Table 4), 59% of participants used the *mode of transportation*, well above the next highest percentage for *types of places mentioned* in the text, which still ranked high (46%). The full results are shown in Table 3. A few characteristics that were not used as often show interesting information. The *number of stops* was not used much, and this shows a clear difference with trajectory research. Determining stops is important in trajectory research for tasks like knowing when people driving in their cars stop at businesses or stop because of traffic. This finding shows that stops are less critical in descriptions of movement. Also, *direction traveled* is used less often. This could be because *direction traveled* is mentioned less frequently in movement statements or because *direction traveled* would require both conceptualizing the movement's path and the difference in the origin and destination. Therefore, it requires two general cognitive steps to conceptualize. *Area covered* and *distance traveled* only require one general cognitive step of visualizing the places. Lastly, our concerns from the pilot survey about *movement acceleration rate* incorrectly ranking high were confirmed because it was selected by only 15% of the participants. We surmise that identifying *movement duration* as an important characteristic helped participants who were possibly confusing *movement acceleration rate* with *time duration*.

**TABLE 4** All other characteristics are shown ranked by the number of times selected as being used and the percentage of times chosen compared to the total possible number of times.

| Characteristic | Count | % of Participants Who Chose |
|---|---|---|
| mode of transportation | 24 | 59 |
| types of places similar – nation, city, ocean, forest | 19 | 46 |
| same or similar places | 18 | 44 |
| distance traveled | 17 | 41 |
| individual or groups of things moving | 16 | 39 |
| movement speed – fast, slow | 14 | 34 |
| things leave and return to the same place | 12 | 29 |
| continuous or non-continuous movement | 11 | 27 |
| time ordered sequence of movements | 10 | 24 |
| number of turns | 10 | 24 |
| movement caused by other incidents | 9 | 22 |
| direction traveled | 9 | 22 |
| number of stops | 7 | 17 |
| movement repeats at intervals - monthly, yearly | 7 | 17 |
| things moving that intersect with other things | 6 | 15 |
| movement acceleration rate | 6 | 15 |
| other | 0 | 0 |

## 4.3   | Categorizing Participants by their Choices

The full survey Likert scale choices were aggregated into two bins of using (those who answered regularly, frequently, or always) vs. not-using (those who answered never or rarely) the characteristic. Figure 6 shows a summary of the participants' use and non-use of the primary characteristics. Here we see that 63% of participants used *time duration*. A similar grouping distribution exists for the *thing moving* and *general movement vs. a specific path*. This shows that *movement duration*, *thing moving*, and *general vs. specific* are more polarizing characteristics than the *area covered*.

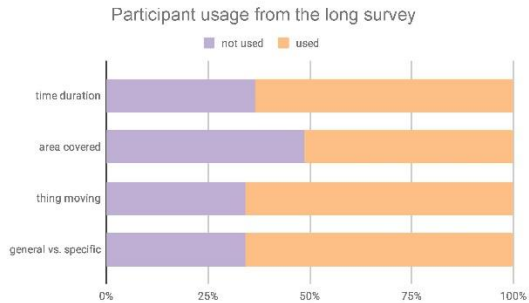

**FIGURE 6**   The percentage of participants responding in the full survey that they use each characteristic (orange) versus the percentage of participants who do not use that characteristic (purple).

It is also possible that the *area covered* can be confused with the *distance traveled*. Depending on the path of movement, these two characteristics could be very similar or very different. If there is one thing moving and moving in roughly a straight path, then the *area covered* and *distance traveled* could be considered the same. In this example, it is still possible that people would conceptualize the *area covered* differently by including the entire area of the political or geographic entities that were traveled through. A second example that shows how the difference between these two characteristics can be confusing is the opposite case from a straight path where the thing moves long distances in circular or irregular paths but stays in the same general (perhaps small) area. In this case, one person could conceptualize the difference between the *area covered* and the distance to be large since the area would be small, but the entire length of the movement is great. Someone could still see these characteristics as similar if they conceptualize distance traveled as the distance from the starting point to the ending point and not consider movement in between.

In the full survey, it seems that participants were loose on their definition of "always" since, for each characteristic, between six to nine participants said they always used it while, in the short survey, no one did choose it for every triad. The most anyone used the *area covered* was 76%, the *time duration* 48%, *thing moving* 83%, and *general vs. specific* 75% of their triads completed.

Next, we categorized the participants into groups of those who used each of the four main characteristics and those who did not use that characteristic. Then, for each characteristic, we compared the responses to the other three characteristics of the participants who used the first characteristic against the other three characteristics of the participants who did not use the first characteristic. This exercise was intended to find overall groups of participants who tended to use two characteristics together or the opposite.

Figure 7a shows the most interesting results of this exercise that participants who did not use the *area covered* by the movement also generally used the *time duration* and the *thing moving* but were split on *general vs. specific*. In contrast, Figure 7b shows those that did not use the *time duration* were split on *area covered* but used *general vs. specific*.

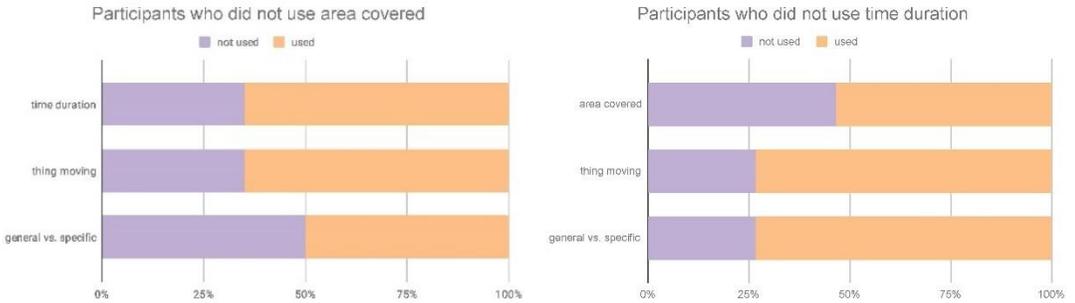

When the *area covered* was not used, these participants did use *time duration* and *thing moving*.

When *time duration* was not used, these participants were split on using *area covered* and used *general vs. specific*.

**FIGURE 7** Participants split by those not using *area covered* and not using *time duration*.

Another interesting result is shown in Figures 8a and 8b where those who used the *thing moving* also used many other characteristics, but most of those who did not use the thing moving did not use the *area covered,* and none of them used *general movement vs. a specific path*. Figure 9a shows that when *general vs. specific* was used, these participants used all other characteristics with the *thing moving, area covered,* and *time duration* from most to least. Figure 9b shows that when participants did not use *general vs. specific,* they also did not use *area covered*. Overall, the participants who used the *thing moving* and *general vs. specific* commonly relied on other characteristics. In addition, those who did not use *general vs. specific* also did not use the other geographic characteristic, *area covered*.

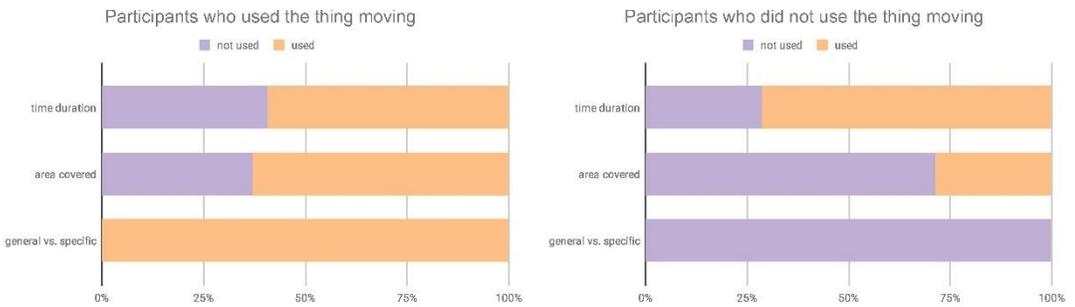

When the *thing moving* was used, these participants *time duration* but they did not use *area covered*.

When *time duration* was not used, these participants did use *area covered* and *general vs. specific*.

**FIGURE 8** When the thing moving was not used, these participants did use general vs. specific.

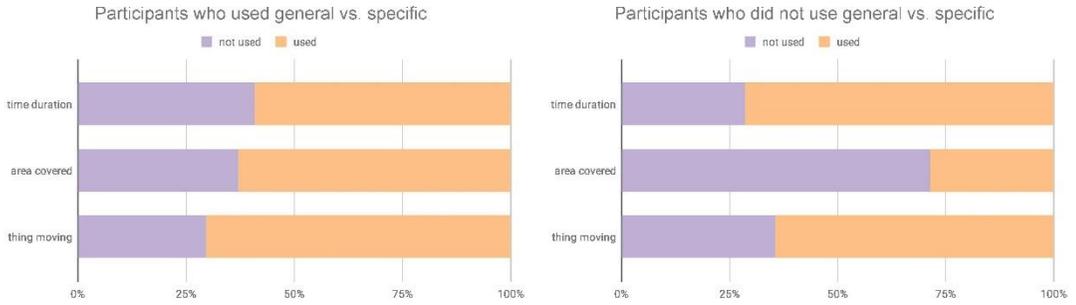

When *general vs. specific* was used, these participants used the *thing moving*, *area covered*, and *time duration* from most to least.

When *general vs. specific* was not used, these participants also did not use the *area covered*.

**FIGURE 9**   When *general vs. specific* was used, these participants used the thing moving, area covered, and time duration from most to least. When *general vs. specific* was not used, these participants also did not use *area covered*.

In summary, in this section, we first show that certain characteristics are used more often (*time duration*) while others are more polarizing (*area covered*) to people. Second, there are some people who tend to use geographic characteristics frequently while others tend not to use them much.

## 4.4 | Participant Choices While Viewing Statements Versus Choices in Hindsight

As mentioned, we conducted our short survey for every triad as the participant was viewing that triad. The full survey was conducted once for the most experienced participants after they were done with their triad tasks and short surveys. Here, we compare their survey answers for the most important characteristics in the short survey while viewing the triads to their full survey answers when thinking back to their completed tasks. To directly compare the many short surveys to the one full survey, we aggregated each participant's short survey answers. We set a cutoff; if the participant chose the characteristics in 15% or more of their triad survey responses, we labeled this participant as using the characteristic. If they said they *used* this characteristic in less than 15% of their triad short survey responses, then we labeled the participant as *not using* the characteristic.

The comparisons made for the short survey exhibited fewer clear results. Perhaps the arbitrary choice we made for a cutoff in the participants who used a characteristic or not based upon the number of times they chose it in the short survey affected the clarity of results. Or perhaps because the lack of clear results in the short survey resulted because participants needed to experience multiple triads before they settled on strategies or realized which factors they were relying upon. We surmise that the short survey is less clear and likely to have produced less consistent results than the full survey because full survey participants had more experience and had likely solidified their judgments on characteristics.

Figure 10 shows that when the survey immediately follows the statements in the short survey, most participants used the *area covered, thing moving,* and *general vs. specific*. They were equally split on *time duration*. The participants used the *area covered* and *time duration* much more often than what they said in the ending full survey (as shown in Figure 6) when thinking back.

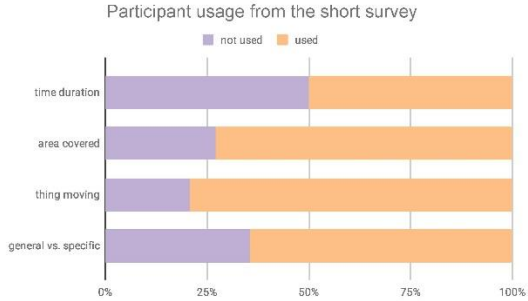

**FIGURE 10** The number of participants responding in the short survey that they use each characteristic versus the number of participants who do not use that characteristic.

When comparing the results of the short survey to the results of the full survey, *time duration* was considered more influential on reflection after doing a set of tasks, and the *area covered* decreased in importance. Possible reasons for this may be that participants did not understand what *time duration* referred to until they had done several tasks; also, as discussed above *area covered* may have been a hard concept to interpret. Despite the hints given about the size of the area covered (local, state, etc.), some participants seemed to interpret it as intended while others seemed to interpret it as meaning that the "*areas covered" was a reference to* descriptions about different parts of the World.

## 5 | RECOMMENDATIONS FOR COMPUTATIONAL METHODS IN ANALYSIS USING MOVEMENT STATEMENTS

It is important to learn about how humans judge differences and similarities between movement described in text documents so that computational and expert systems can improve analysis and visualization. For example, just as recommender systems benefit from knowing why sentences are different or similar, in order to recommend similar products, we can find similar movement statements, summarize movements in statements, and find patterns in movement statements by knowing why they are similar and different. In general, computational techniques that summarize large amounts of text (Aliguliyev, 2009), text mining, and IR benefit from knowing why sentences are semantically similar (Achananuparp et al., 2008).

The full survey and short survey results suggest multiple computational methods that researchers can use to aid analysis. First, the responses and analysis above hint at a long-standing conclusion in the field of cognitive geography that some people simply think more spatially than others (Liben and Downs, 1993; Ishikawa, 2016; Montello et al., 1999). Therefore, some people focused on the *area covered* and *general movement vs. a specific path,* while others focused on the *time duration* and *thing moving*. Based on these results, one recommendation for future research and analysis would be: any visualization of results should have a map view and other views of the data, such as a time chart. Anselin et al. (2010) show how providing multiple views of data is important towards broad spatial data analysis. Pezanowski et al. (2017); MacEachren et al. (2011); Karimzadeh et al. (2020) utilize multiple views in geovisual analytics interfaces to show geographic data in text documents. Other research into broader visual analytics topics also stresses the importance of multiple views (Andrienko et al., 2020). Our research provides formal evidence to support the importance of multiple views of data in a visual analytics system to analyze movement described in text documents.

The full survey also showed that transportation mode is important, and this information can be utilized computationally in multiple ways. Named entity recognition (NER) models can be trained to identify vehicles of transportation in text. Or NLP can be used to look for linguistic phrases related to different modes of travel depending on the focus of the thing moving. Even simpler methods like string pattern matching would likely detect a large portion of references to modes of travel.

A final recommendation is that GIR can help focus on the types of places and same or similar places. Most GIR systems provide metadata about places mentioned in the text. This metadata, such as whether the places are physical features, countries, or cities, can be used in the analysis to group movement (Yang et al., 2020). In addition, our finding that *types of places* are an important characteristic to movement statements is valuable to research into GIR systems for geoparsing and geocoding place mentions. Karimzadeh et al. (2019) already showed that the type of place is an important attribute for ranking geocoding results suggestions about which place is the correct place for places with common names. Our findings support the previous findings that place type is valuable information for resolving place mentions in text to their intended location. Our research, together with prior results using co-occurring place mentions in text to improve GIR place name resolution (Karimzadeh et al., 2019), extends these findings by showing the characteristics' importance when place mentions are connected by implied movement. Many of the other chosen characteristics can also be analyzed in more depth with additional language and IR techniques.

# 6 | SUMMARY OF FINDINGS AND CONCLUSIONS

Geographic movement described in text documents contains a wealth of geographic and contextual information about the movement of people, wildlife, goods, and much more. Unlike explicit movement in precise movement trajectories, movement described in text needs to be identified and extracted before analysis. We need to understand better how humans differentiate the movement described so that future research and computational analysis can focus on these characteristics. To investigate characteristics that humans may use to differentiate statements, we conducted a statement comparison experiment with human participants who were asked to choose the most different statement out of three. Also, the participants were asked what characteristics they used in making their choice about statements that are most different.

This research makes three overall contributions to improve our understanding of movement in text. First, we show that interpreting movement described in statements is a difficult task even for humans. Many linguistic features make the task difficult. Second, the findings answer our research question: There are identifiable characteristics of descriptions of geographic movement that people use to differentiate the described movement. The most important characteristics were determined to be the movement's time duration, the type of thing that is moving as described in the statement, and geographic area covered by the movement described in the statement. From our own experience cleaning the statements before doing the pilot test and again creating the final text, we also identified the importance of *general movement versus a specific path*, which was confirmed in the full experiment. This characteristic was deemed important by the participants but slightly less so than the other three. That may be because it is less important or because the characteristic referred to by *general vs. specific* was more abstract and less clear to participants. Other characteristics were also shown to be important to people, and these can be used in future research as a framework for analyzing geographic movement found in written text. Third, based on our extensive experience with movement statements and the results of our experiments, we make recommendations for both scientists who intend to do research with geographic movement described in text and for anyone wanting to apply analysis on this unused dataset in practice. In Section 5, we list multiple computational techniques that can benefit analysis and multiple recommendations for visual analytics system configurations.

In addition to these findings, we offer three observations. First, the results show that the type of thing moving is a natural way to differentiate statements. This aspect poses challenges to unsupervised clustering algorithms that tend to use this characteristic. In some cases, the type of thing moving may not need to be separated from the geographic movement during analysis. An example may be applications of deep learning to classify movement types where the machine learning model will decide what is important or not. An example where it would be important to separate the kind of movement from the type of thing moving is analysis to find different movement patterns among the same type of thing. Second, there is a distinction between people who focus on geographic characteristics to differentiate movement statements while others focus on non-geographic characteristics. Third, other essential characteristics can differentiate movement statements like transportation mode and similar types of places mentioned.

Considering these findings, an overall recommendation for researchers interested in using movement statements in their analysis is that they should combine computational methods with human sensemaking of the computational predictions. We showed that interpreting and understanding movement statements is challenging even for a human on a small number of statements. Movement statements often include vague geographic references, are complicated by references to multiple things moving at different times or groups of things, contain slang or jargon, and sometimes lack significant context. Given all these factors, it makes sense that much computational analysis and machine learning predictions are required, and these techniques will be imprecise. Therefore, a human-in-the-loop approach using techniques like geovisual analytics that allow humans to understand and correct predictions is vital. The first step towards such a system is the GeoMovement web application (Pezanowski, 2021), where movement statements can be explored using a human-in-the-loop sensemaking approach from the application of computational techniques.

The recommendations from our findings in this research can be applied to applications designed to analyze movement described in text. Because analysis of movement described in text is likely to involve combining multiple imprecise computational methods, a geovisual analytics sensemaking application would be a good fit for the information's human synthesis. A sensemaking application of movement described in text should have multiple views of the data that utilize a map view for users more focused on the geographic aspect of the statements and non-geographic views for those who focus less on geographic aspects (e.g., those that emphasize the temporal components of movement). Geographic movement statements are diverse in their spatial and temporal characteristics, topics, linguistic characteristics, and movement patterns. Therefore, they should be analyzed using many methods that combine computational techniques with human analysis.

Our research and future research into descriptions of geographic movement not only will help uncover knowledge about things moving but can also improve other related research. The research can:

- Advance GIR by identifying essential characteristics or features in recognizing and resolving place names
- Progress research in spatial linguistics
- Be used as input to research conversational systems that are understandable and realistic with directions and geographic relationships.

Future research should also go deeper into why people decided that specific characteristics were important and what caused the people to differ in their interpretations about statement similarity and difference, as well as about the characteristics that are key to making such interpretations. For example, how do place names that are unfamiliar to the participants affect their interpretation of the movement? And how does the combination of short paths through places that are large in area affect people's interpretation of the route? Also, does people's prior knowledge or perceptions of the movement by the type of thing affect their interpretation of the movement described. We anticipate our efforts to identify essential characteristics of movement statements will prompt future research like how the starting point frameworks about geographic trajectories that we used did.

The authors submitted the study to their institutions' Institutional Review Board (IRB) for the experiments using human participants, and it was approved as "exempt."